 \documentclass[final,3p,times,twocolumn]{elsarticle}
\usepackage{amssymb}
\usepackage{hyperref}
\usepackage{graphicx}
\usepackage{subcaption}
\usepackage{lscape}
\usepackage{amsmath}
\usepackage{multirow}
\usepackage[figuresright]{rotating}
\usepackage{lineno}
\usepackage{algorithm} 
\usepackage{algpseudocode} 
\usepackage{comment}
\usepackage{lineno}
\usepackage{newtxtext,newtxmath} 

\usepackage{natbib}
\setcitestyle{authoryear,open={(},close={)},citesep={;}}

\begin{document}

\begin{frontmatter}



\title{High-Precision Fruit Localization Using Active Laser-Camera Scanning: \\Robust Laser Line Extraction for 2D-3D Transformation}

\author[label1]{Pengyu Chu}
\author[label2]{Zhaojian Li}
\author[label2]{Kaixiang Zhang}
\author[label1]{Kyle Lammers}
\author[label3]{Renfu Lu}

\address[label1]{Department of Electrical and Computer Engineering, Michigan State University, East Lansing, MI 48824, USA}
\address[label2]{Department of Mechanical Engineering, Michigan State University, East Lansing, MI 48824, USA}
\address[label3]{United States Department of Agriculture, Agricultural Research Service, East Lansing, MI 48824, USA}

\begin{abstract}
Recent advancements in deep learning-based approaches have led to remarkable progress in fruit detection, enabling robust fruit identification in complex environments. However, much less progress has been made on fruit 3D localization, which is equally crucial for robotic harvesting. Complex fruit shape/orientation, fruit clustering, varying lighting conditions, and occlusions by leaves and branches have greatly restricted existing sensors from achieving accurate fruit localization in the natural orchard environment. In this paper, we report on the design of a novel localization technique, called Active Laser-Camera Scanning (ALACS), to achieve accurate and robust fruit 3D localization. The ALACS hardware setup comprises a red line laser, an RGB color camera, a linear motion slide, and an external RGB-D camera. Leveraging the principles of dynamic-targeting laser-triangulation, ALACS enables precise transformation of the projected 2D laser line from the surface of apples to the 3D positions. To facilitate laser pattern acquisitions, a Laser Line Extraction (LLE) method is proposed for robust and high-precision feature extraction on apples. Comprehensive evaluations of LLE demonstrated its ability to extract precise patterns under variable lighting and occlusion conditions. The ALACS system achieved average apple localization accuracies of $6.9$ - $11.2$ mm at distances ranging from $1.0$ m to $1.6$ m, compared to $21.5$ mm by a commercial RealSense RGB-D camera, in an indoor experiment. Orchard evaluations demonstrated that ALACS has achieved a $95$\% fruit detachment rate versus a $71$\% rate by the RealSense camera. By overcoming the challenges of apple 3D localization, this research contributes to the advancement of robotic fruit harvesting technology.

\end{abstract}


\end{frontmatter}
% Start line numbering here if you want
% \linenumbers

\section{Introduction}
\label{sec:intro}

Robotic technology has attracted much research attention in recent years due to its potential to revolutionize the agricultural industry. With the rising labor cost and increasing demand for fruit production~\citep{Apple_stat}, manual harvesting has become a major issue for the sustainable production of fruit in the U.S. and other developed economies. The development of robotic systems~\citep{De2011, DeASABE2015, SilwalJFR2017, Mehta2014CEA, Xiong2018CEA, Williams2019BE, Zhang2022IROS} specifically designed for fruit harvesting has emerged as a promising solution to address the labor and production cost challenges. Robotic fruit harvesting has the potential to increase harvest efficiency, reduce the industry’s reliance on manual labor, promote sustainable farming practices, and enhance the overall profitability of the agricultural sector~\citep{raj2022precision}. A typical harvesting robot is mainly composed of a manipulator, an end-effector, and a fruit perception system. The fruit perception system plays a key role in fruit harvesting; it performs the tasks of fruit detection and three-dimensional (3D) localization~\citep{fu2020application}, so that the robot’s manipulator and end-effector can carry out effective and efficient harvesting tasks.

Thanks to the great success on object detection and semantic image segmentation, deep learning-based fruit detection approaches~\citep{sa2016deepfruits, bargoti2017image} have demonstrated promising performance by overcoming challenges in the natural orchard environment, such as illumination and appearance variances, noisy background, and occlusions. Promising performance has been achieved in various studies \citep{chu2021deep,sa2016deepfruits,rahnemoonfar2017deep,fennimore2019robotic, ahmad2021performance, Jocher_YOLO_by_Ultralytics_2023}.
In particular,  the  Occluder-Occludee Relational Network (O2RNet)~\citep{chu2023o2rnet} developed in our prior work has outperformed  other state-of-the-art models~\citep{oerke2006crop, norsworthy2012reducing, fennimore2019robotic, ahmad2021performance, Jocher_YOLO_by_Ultralytics_2023} with a higher F1-score of $0.873$ on our customized dataset by explicitly addressing partial occlusions of clustered apples.

Three-dimensional localization is the other crucial aspect of fruit perception. Accurate and reliable localization of objects such as fruits in the 3D space is essential for automated agricultural systems to optimize crop management and harvesting strategies, ultimately improving productivity, efficiency, and sustainability~\citep{ge2019fruit}.  Several types of 3D sensors are currently available, including Time-of-Flight (ToF) cameras, LiDAR (Light Detection and Ranging), stereo-vision cameras, and structure light systems. Specifically, ToF cameras \citep{lanzisera2011radio} measure depth by emitting a light signal (usually IR) and measuring the time it takes for the signal to bounce back to the sensor. This time difference allows the camera to calculate the distance to objects in the scene. ToF cameras, such as the PMD CamBoard pico \citep{bahnsen20213d} and Microsoft Kinectv2 \citep{neupane2021evaluation}, are known for their accuracy and high-resolution depth data. LiDAR systems \citep{raj2020survey} use lasers to send out light pulses and measure the time it takes for the pulses to return after reflecting off objects. This time difference is used to calculate distances and generate a 3D point cloud of the scene. LiDAR systems can provide accurate depth data but are often more expensive than other methods and often have limited spatial resolution. Stereo vision systems, on the other hand, use two cameras, typically placed side-by-side at a known distance apart, to capture images of the same scene. By comparing the images and identifying corresponding points in each image, depth information can be estimated using triangulation. Stereo vision systems \citep{lazaros2008review} often require more computational power and can be sensitive to lighting conditions, but they do not rely on active IR illumination. In contrast, structured light systems \citep{trtik2019methods}  project a known pattern of light (usually infrared) onto the scene and then capture the deformed pattern with a camera. The deformation of the pattern allows the system to reconstruct the 3D geometry of the scene. It usually works well in low-light conditions (since it uses active illumination) but it is sensitive to ambient light and surface properties (e.g., reflectivity, transparency).

Over the years, numerous techniques have been attempted for fruit 3D localization based on the aforementioned sensors and advanced computer vision methods~\citep{mehta2016multi, habibie2017fruit, yu2022mature, liu2023orb}. 
Specifically, \cite{mehta2017multiple} employed a stereo camera system in conjunction with a tailored fruit-matching algorithm, reporting a localization error of approximately $11$ mm in the simplified indoor environment. However, outdoor agricultural environments are significantly more challenging,  with exposure to a wide spectrum of lighting conditions and complex tree and fruit structures, which can cause major issues to its fruit matching techniques as they can significantly influence the visual characteristics and discernibility of fruits captured within the images or point clouds.
\cite{andriyanov2022intelligent} utilized a commercial device, RealSense RGB-D camera, to obtain the positions of apples and reported a localization error of around $9.5$ mm in an ideal indoor environment. However, due to the low resolution of the projector in the depth camera, RGB-D cameras have to interpolate based on partial depth measurements. Unlike plane surfaces, fruits can exhibit a wide range of shapes, sizes, colors, and textures, making it difficult to develop a one-size-fits-all approach to 3D fruit localization. 
To overcome complex fruit morphology, \cite{silwal2017design} utilized a global time-of-flight camera to obtain fruit point cloud by removing the background, in order to accurately localize each point on the fruit. However, in real-world orchard scenarios, fruits are often surrounded by leaves, branches, and other fruits, which can create occlusions and clusters in the images or point clouds. These issues make it difficult to accurately identify and localize the fruits, particularly when they are partially or fully occluded.

In this study, we address the above challenges by designing and developing a novel  Active
Laser-Camera Scanner (ALACS) system. Specifically, an RGB camera is integrated with a line laser to achieve robust and accurate localization using the triangulation principle. We propose a feature-matching algorithm, called Laser Line Extraction (LLE), to help ALACS transform the 2D fruit positions to 3D fruit positions, thus achieving accurate mapping even under complex fruit morphology, variable lighting conditions and occlusions. This research is expected to provide a valuable reference for future research on developing fruit 3D localization systems in fruit harvesting. The main contributions of this paper are highlighted as follows:

\begin{enumerate}
\item A novel Active Laser-Camera Scanning system, consisting of a red line laser, an RGB-D camera, and a linear motion slide, is designed and developed for accurate fruit 3D localization.
\item A Laser Line Extraction (LLE) algorithm is proposed and implemented  for robust feature matching to enable stable 2D-3D transformation for ALACS.
\item System evaluation and validation of the ALACS are performed indoors and outdoors in comparison with a conventional 3D sensing technique, i.e.,  Intel RealSense D435i RGB-Depth camera. 
\end{enumerate}

The remainder of the paper is organized as follows. The ALACS system is first introduced in Section~\ref{sec:alacs} and the LLE algorithm applied to red apples is presented in Section~\ref{sec:lle}. Section~\ref{sec:evaluation} presents a comprehensive system evaluation for ALACS both indoors and outdoors. Finally, concluding remarks are presented in Section \ref{sec:conclu}.

\section{Active Laser-Camera Scanning (ALACS)}
\label{sec:alacs}

\begin{figure}[!h]
	\centering
	\includegraphics[width=0.37\textwidth]{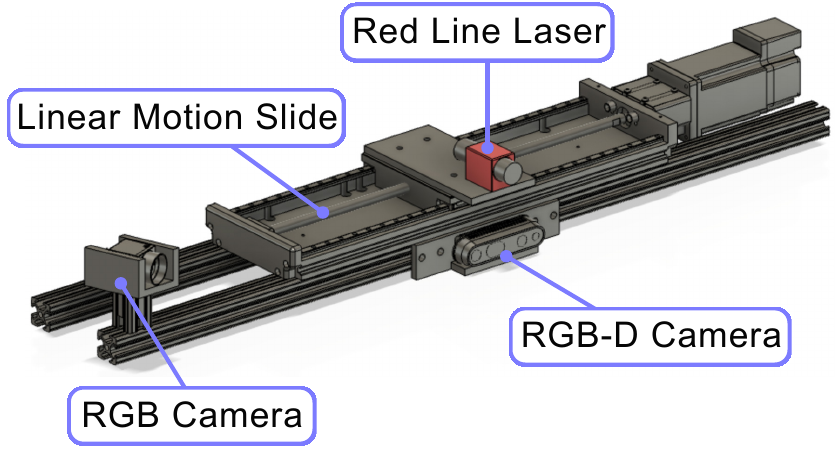}
	\caption{CAD model of ALACS.}
	\label{fig:alacs}
\end{figure}

The ALACS is designed to provide accurate 3D localization of apples in orchards by combining the advantages of a depth camera and laser scanning. The major hardware components include a red line laser (Laserglow Technologies, North York, ON, Canada), a FLIR RGB camera (Teledyne FLIR, Wilsonville, OR, USA), a linear motion slide, and an Intel RealSense D435i RGB-D camera, which is used for providing rough initial global estimates of fruits. As shown in Figure~\ref{fig:alacs}, the RGB-D camera is mounted on a horizontal frame that is above the manipulator to provide a global view of the scene and initialize the laser position. The line laser is mounted on the linear motion slide that enables the laser to move horizontally with a full stroke of $20$ cm. Meanwhile, the FLIR RGB camera is installed at the left end of the linear motion slide with a relative angle to the laser to capture laser patterns on apples. The hardware configuration of the ALACS is designed to facilitate depth measurements based on the principle of laser triangulation \citep{dorsch1994laser}. Specifically, the laser triangulation-based technique is a classical high-precision localization scheme that captures depth measurements by pairing a laser illumination source with a camera. It is worth noting that with the conventional laser triangulation sensors, the relative position and pose between the laser and the camera is fixed (i.e., both of them are static or moving simultaneously), whereas in ALACS the camera is fixed while the laser position is actively adjusted with the linear motion slide to seek the target fruit (see subsequent discussions for more details). Specifically, ALACS performs fruit 3D localization in three steps: laser scanning, target position determination, and 2D-3D position transformation.

\begin{figure}[!h]
	\centering
	\includegraphics[width=0.45\textwidth]{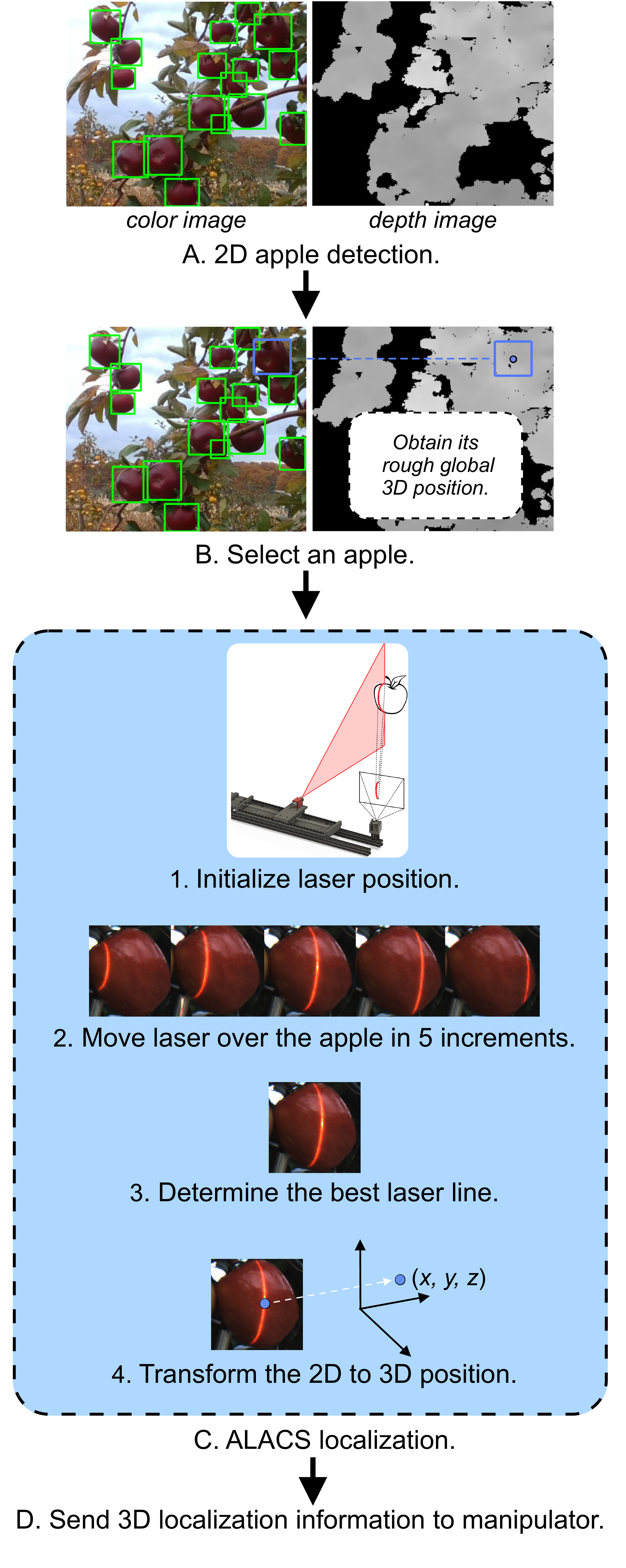}
	\caption{Schematics of the ALACS workflow: In step A, the RGB-D camera provides an initial global view of detected apples. In step B, a target apple is determined based on a planning strategy and its rough 3D location is sent to ALACS. In step C, the fruit is scanned and the high-precision position is obtained. Finally in step D, the 3D localization information is sent to the manipulator for fruit picking.}
	\label{fig:workflow}
\end{figure}

%\subsection{Laser scanning}

\begin{figure}[!h]
	\centering
	\includegraphics[width=0.5\textwidth]{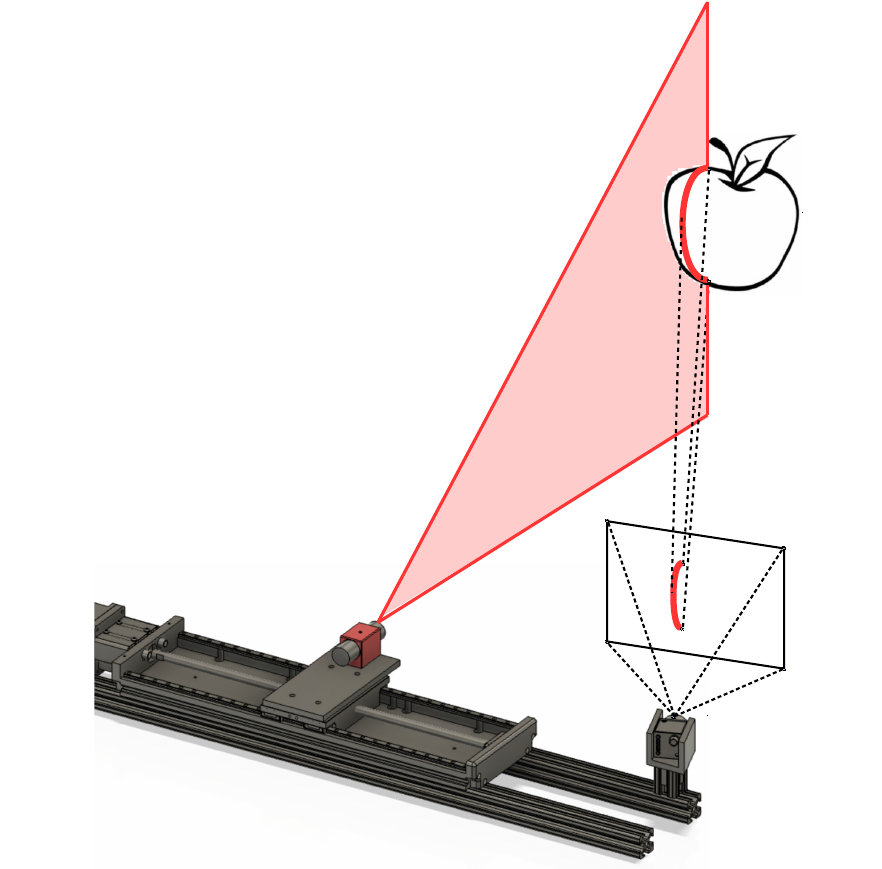}
	\caption{Schematic of  laser scanning on a target fruit in ALACS to obtain one laser line from the fruit.}
	\label{fig:scanning}
\end{figure}

As shown in Figure~\ref{fig:workflow}, the first step in the ALACS workflow is to capture a global view of the scene using the RGB-D camera. This camera provides both color and depth information, which can be used to segment and identify bounding boxes containing apples based on an apple detection approach (see~\citep{chu2023o2rnet}). Then, a rough 3D position of each detected apple is estimated. Based on a planning strategy~\citep{ZhangJFR2013}, ALACS selects a target apple and uses the rough 3D position provided by RealSense D435i to guide the laser to the initialized position related to the target apple. The laser is then moved horizontally from the left to the right side of the apple in five $2$-cm increments; it illuminates the target apple and creates visible laser lines on the surface of the fruit (see Figure~\ref{fig:scanning}). 

During the laser scanning process, at each stop an image is obtained at each stop with the illuminated apple being captured by a high-resolution FLIR camera. ALACS uses a laser line extraction algorithm (see Section~\ref{sec:lle}) to extract laser patterns on the target apple for each stop. These candidates resulting from the five different laser projections would cover the target apple. The most reliable candidate is then selected to determine the apple's centroid position based on a confidence evaluation for each candidate, which is calculated using two key factors: the distance to the estimated center and the number of extracted laser line pixels.

\noindent \textit{Distance to the Estimated Center}: The distance factor in the confidence calculation quantifies how close the candidate laser line is to the apple's estimated center (obtained from our apple detection algorithm \citep{chu2023o2rnet}). Candidates that are closer to the center are considered more reliable and are assigned higher confidence scores. This distance factor $\Delta d$ is calculated using the Euclidean distance between the candidate's position and the estimated center.

\noindent \textit{Number of Extracted Laser Line Pixels}: The second factor contributing to the confidence calculation is the number of extracted laser line pixels, $\mathcal{N}$, for each candidate. Candidates with more extracted laser line pixels are considered to provide a more complete representation of the apple's surface geometry and are thus deemed more reliable. Consequently, these candidates are assigned higher confidence scores. Then the confidence $\mathcal{P}$ is calculated by $\mathcal{P} = \omega_1 \cdot \mathcal{N} - \omega_2 \cdot \Delta d$, where $\omega_1$ and $\omega_2$ are weights for these two factors and are obtained through cross-validation. 

After selecting the most reliable candidate based on the calculated confidence scores, the 2D position of the center of this laser line is obtained and the apple's center position is determined using the laser triangulation scheme \citep{dorsch1994laser}. The basic idea of this technique is to capture depth measurements by pairing a laser illumination source with a camera. Both the laser beam and the camera are aimed at the target object, and based on the extrinsic parameters between the laser source and the camera sensor, the depth information can be computed using trigonometry. The transformation rule for 2D position $(u_i, v_i)$ to 3D position $(x_i, y_i, z_i)$ follows  

\begin{equation} \label{eq:transformation}
	\begin{aligned}
 	    z_{i} &= \frac{L}{\sin(\alpha) - u_{i}\cos(\alpha) - v_{i}\tan(\beta)}, \\
		x_{i} &= \frac{Lu_{i}}{\sin(\alpha) - {u}_{i}\cos(\alpha) - {v}_{i}\tan(\beta)}, \\
		y_{i} &= \frac{L{v}_{i}}{\sin(\alpha) - {u}_{i}\cos(\alpha) - {v}_{i}\tan(\beta)},
	\end{aligned}
\end{equation}

\noindent where extrinsic parameters $L$, $\alpha$, and $\beta$ are, respectively, the baseline (i.e., the distance between the camera and the line laser), horizontal angle, and vertical angle between the laser illumination source and the camera. The details of parameter estimation are discussed in ~\citep{Zhang2023CEA}.
Therefore, ALACS finally localizes the target apple. After the apple has been picked, the process repeats for the next target fruit.

\section{Laser Line Extraction (LLE)}
\label{sec:lle}
In this section, we present more details on the laser line extraction steps that are of paramount importance in the ALACS system since it serves as the crucial link between the laser scanning process and the final triangulation-based localization. In ALACS, images of the illuminated apple are captured using a high-resolution camera. These images are then processed to extract the laser lines on the apple's surface. Extracting accurate and well-defined laser lines from the captured images provides essential geometric information about the apple's surface. Furthermore, effective laser line extraction techniques can help mitigate the impact of noise, occlusions, and illumination variations, which can significantly improve the overall performance and reliability of the ALACS method. We next discuss the specific algorithms and techniques employed for laser line extraction and their role in enhancing the accuracy of the ALACS-based apple localization process.

\begin{figure*}[!h]
	\centering
	\includegraphics[width=0.9\textwidth]{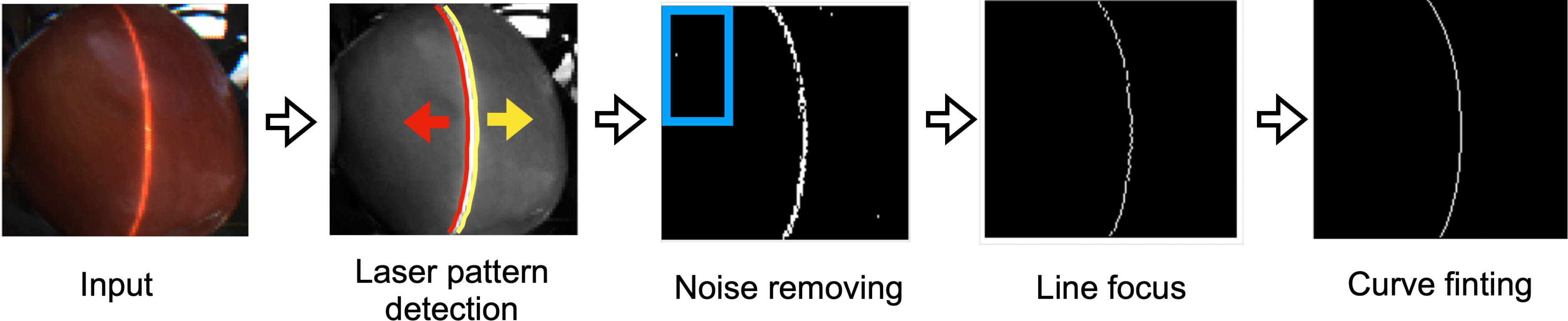}
	\caption{Schematic of the  laser line extraction (LLE) workflow.}
	\label{fig:lle}
\end{figure*}

\begin{figure}[!h]
	\centering
	\includegraphics[width=0.45\textwidth]{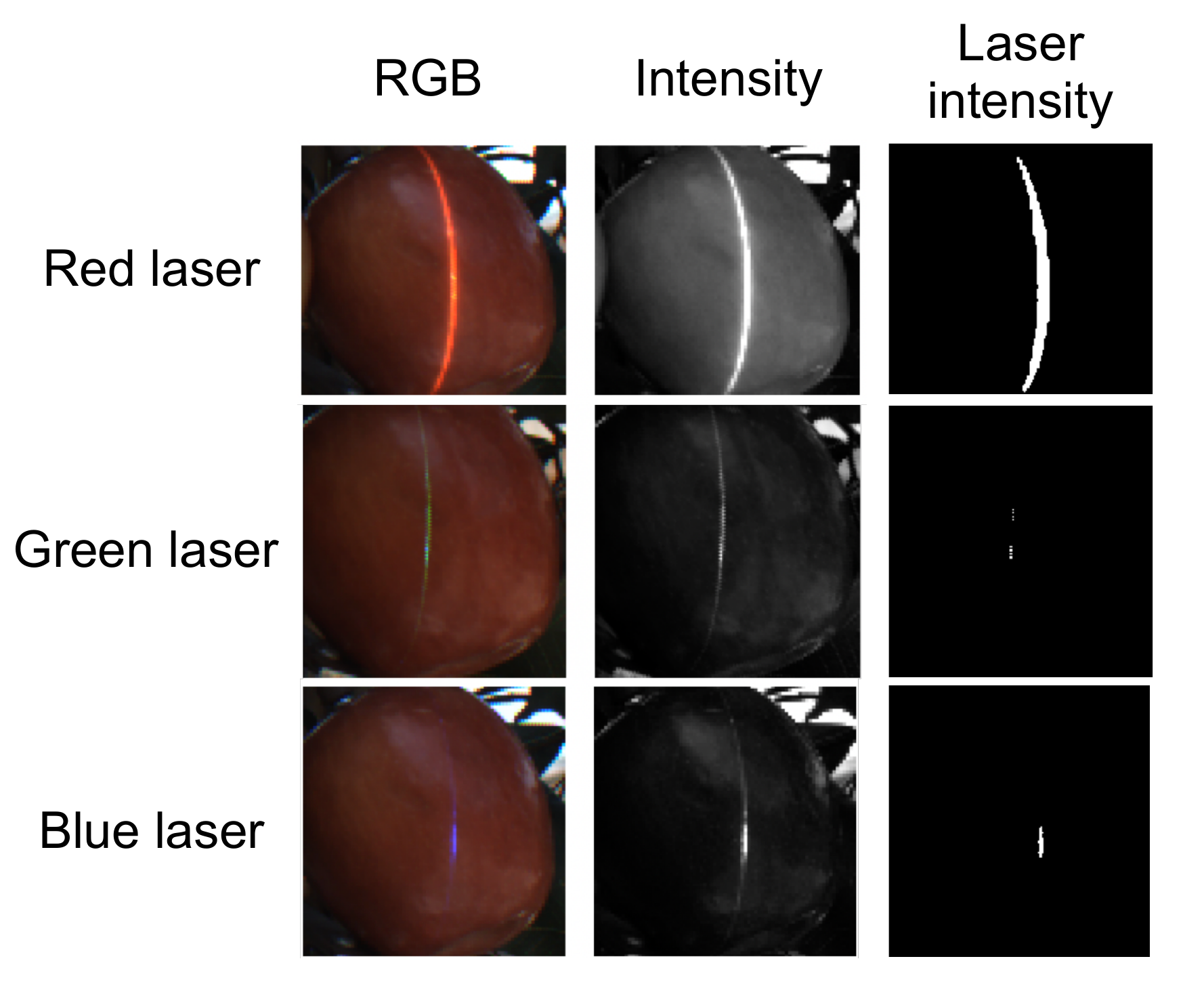}
	\caption{Comparison of laser projections on a red apple by a red laser ($635$ nm), green laser ($515$ nm) and blue laser ($447$ nm). Laser intensity is obtained using a thresholding of $140$. }
	\label{fig:laser_color}
\end{figure}

We have chosen a red laser as the domain laser color, as opposed to blue or green lasers. This choice was made based on our preliminary evaluations, which found  that red laser lines appear more intense and distinct in the captured images (see Figure~\ref{fig:laser_color} for comparisons among three lasers of different colors), thus facilitating more effective extraction of laser lines. More specifically, LLE is designed with 4 steps: laser pattern detection, noise removal, line focus, and curve fitting, as illustrated in Figure~\ref{fig:lle}.

\begin{algorithm}[!t]
	\caption{bRCE laser pattern detection}
    \label{alg:bRCE}
    \textbf{Input:} $\mathcal{I}$: $n\times m \times 3$   RGB image matrix of the selected apple. \\
    \textbf{Output:} $\mathcal{G}$: $n\times m$ laser prediction matrix. \\
    \textbf{Params:} $step$: gradient step,  $th$: gradient threshold. 
    % \textbf{Params:} $\omega$: gradient step, \\
    % \hspace*{3.7em} $\Delta$: gradient threshold. \\
	\begin{algorithmic}
        % \State $\mathcal{I}_1, \mathcal{I}_2 = Zeros(n, m)$ \\
        % $\mathcal{I}_1[:, :-step] = \mathcal{I}[:, step:]$ \\
        % $\mathcal{I}_2[:, step:] = \mathcal{I}[:, :-step]$ \\
		% \State Obtain $\mathcal{I}_{-step}$ by moving $\mathcal{I}$ left with a $step$ \\
        
  %       Obtain $\mathcal{I}_{+step}$ by moving $\mathcal{I}$ right with a $step$ \\
        \State $\mathcal{R}$ = red channel of $\mathcal{I}$
        \State
        $\mathcal{G}_1 = \mathcal{R}[:, :-2\cdot step] - \mathcal{R}[:, step:-step]$ \\
        $\mathcal{G}_2 = \mathcal{R}[:, 2\cdot step:] - \mathcal{R}[:, step:-step]$ \\
        /*$\mathcal{G}_1, \mathcal{G}_2$ are two horizontal gradients of $\mathcal{I}$.*/
    	\For{$\mathcal{G}_1, \mathcal{G}_2$}
             %    \If{$\mathcal{G}_i \leq th$}
             %    \State $\mathcal{G}_i = 0$
    	        % \EndIf
             \State $\mathcal{G}_i[\mathcal{G}_i \leq th] = 0$
    	\EndFor
		
        \State {$\mathcal{G} = \mathcal{G}_1 \circ \mathcal{G}_2$}, \algorithmiccomment{$\circ$ is element-wise product}
        \State $\mathcal{G}[\mathcal{G} > 0] = 1$

        % \For{$i=1,2,3 \dots n$}
        %     \For{$j=1,2,3 \dots m$}
        %     \If{$\mathcal{G}_{1}(i, j) \leq th$}
        %     \State $\mathcal{G}_{1}(i, j) = 0$,
        %     \EndIf
        %     \If{$\mathcal{G}_{2}(i, j) \leq th$}
        %     \State $\mathcal{G}_{2}(i, j) = 0$,
        %     \EndIf
        %     \State $\mathcal{G}(i, j) = \mathcal{G}_1(i, j) \cdot \mathcal{G}_2(i, j)$
        %     \If{$\mathcal{G}(i, j) > 0$}
        %     \State $\mathcal{G}(i, j) = 1$
        %     \EndIf
        %     \EndFor
        % \EndFor
 
    \end{algorithmic}
\end{algorithm}

\begin{figure}[!h]
	\centering
	\includegraphics[width=0.45\textwidth]{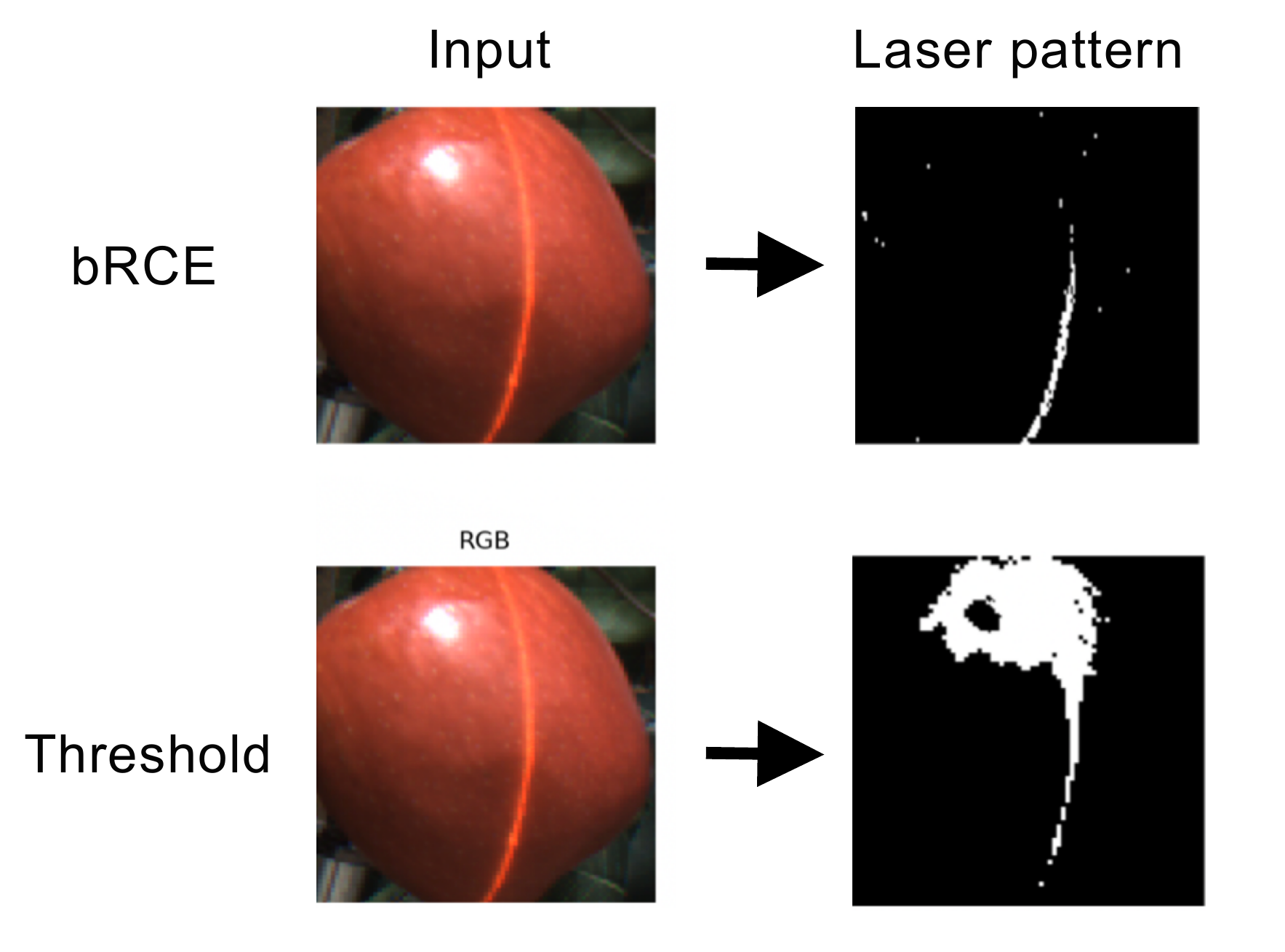}
	\caption{Laser pattern detection comparison between bidirectional Relative Color Enhancement (bRCE) with step$=8$, th$=40$ and thresholding with th$=220$ under over-exposure situations. }
	\label{fig:brce}
\end{figure}

\noindent \textit{Laser Pattern Detection}: Based on the utilized laser line pattern, various image processing techniques, such as edge detection, filtering, and thresholding, can be employed to identify and isolate the laser lines in the captured images. Thresholding is the simplest way to extract highlighted patterns but it is usually affected by strong external lighting like sunlight. Edge detection using Sobel or Robert kernel~\citep{dharampal2015methods} can find line patterns accurately but they need more computation time compared to thresholding. Hence, we designed a novel algorithm, called bidirectional Relative Color Enhancement (bRCE), to detect the laser line pattern from the selected apple image $\mathcal{I}$ efficiently. The bRCE (see Algorithm~\ref{alg:bRCE}) gets bi-directional horizontal gradient matrix$\mathcal{G}$, which is calculated through shifted differences of the image in both left-to-right and right-to-left directions. The bRCE can effectively highlight the boundaries of the laser lines, help enhance the contrast of the laser lines and make them more distinguishable from the background. The bRCE outperforms the thresholding-based method especially in over-exposure situations, as shown by an example in Figure~\ref{fig:brce}.

\begin{figure*}[!h]
	\centering
    \includegraphics[width=0.8\textwidth]{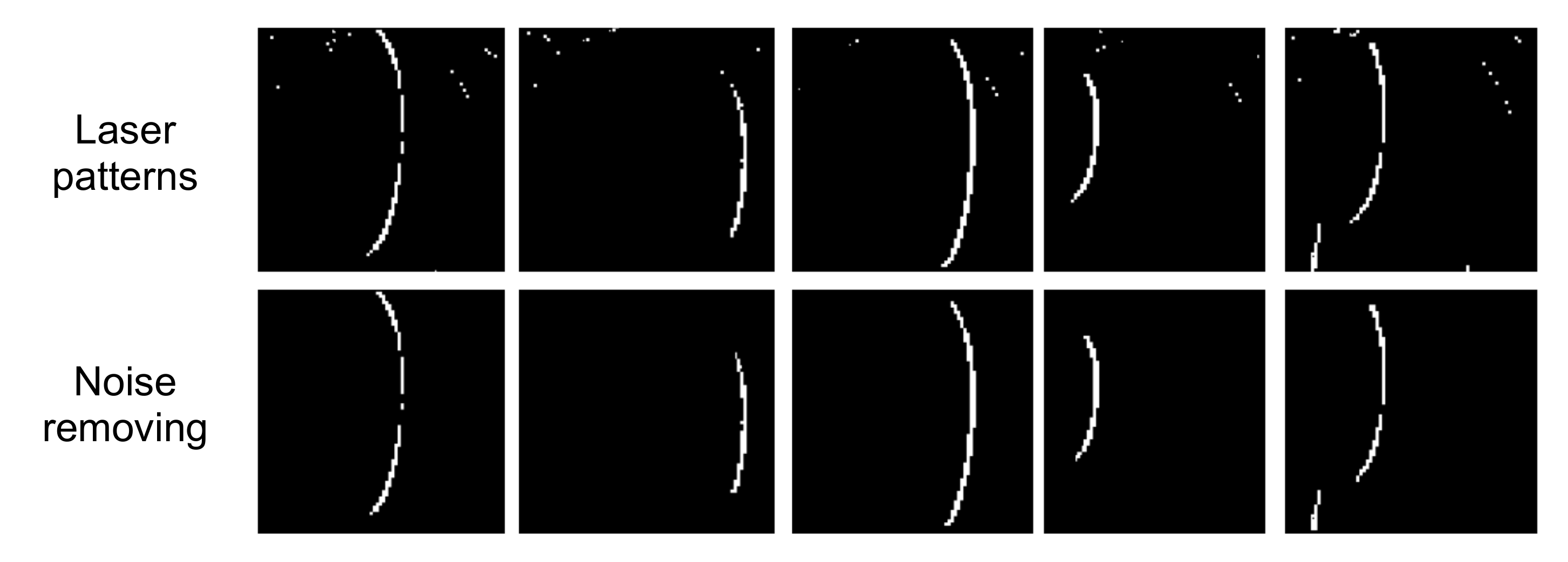}
	\caption{Extracted laser patterns with outliers (noises) (first row) and resultant laser patterns after noise removal with the use of a threshold of $\theta=8$ and the partition size of $\gamma=3$ (second row).}
	\label{fig:noise_remove}
\end{figure*}

\noindent \textit{Noise removing}: The laser line patterns are usually highlighted with small outliers, as shown by examples in Figure~\ref{fig:noise_remove}, since the over-exposure part caused by strong sunlight can interfere with the laser line. To address the discontinuity of the outliers, we employ a sliding window counting method to remove these outliers. As shown in Algorithm~\ref{alg:noise},  a predefined window with size $\mathcal{W}$ is slid horizontally across the image row by row based on the partition size $\gamma$, and the number of strong curve pixels within the window is counted. If the count exceeds a pre-specified threshold $\theta$, the pixels within the window are considered part of the laser line; otherwise, they are categorized as noises and discarded. This filtering process helps retain only the most prominent laser lines $\mathcal{C}$ in the image while eliminating undesired artifacts and noises.

\begin{algorithm}[!t]
	\caption{Noise removing}
    \label{alg:noise}
    % \hspace*{\algorithmicindent}
    \textbf{Input:}   $\mathcal{G}$: $n\times m$ laser prediction matrix (from Algo. 1). \\
    \textbf{Output:} $\mathcal{C}$:  $n\times m$ noise-free laser matrix. \\
    \textbf{Params:} $w_G, h_G$: width and height of $\mathcal{G}$,\\
    \hspace*{3.7em} $\theta$: threshold of noise,  $\gamma$: partition size of $\mathcal{G}$. 
	\begin{algorithmic}
        
		\State /* $\mathcal{W}_{x, y, w, h}$ is the sliding window with the left-top position $(x, y)$,  the width $w$ and the height $h$. */\\
        % $w, h$ and $\gamma$ are hyper parameters\\
        $w, h=\frac{1}{4} w_G, \frac{1}{2} h_G$ \\
        s.t. $ 0 < \theta \leq w\times h $, $ \frac{h}{w} > 1$ \\
        $\mathcal{C} = \mathcal{G}$ \\
        $u = \frac{w_G-w}{\gamma}$ \\
        $v = \frac{h_G}{h}$ 
    	\For{$i=1,2,3 \dots u$}
            \For{$j=1,2,3 \dots v$}
            \State $x = (i-1)\cdot \gamma$,
            \State $y = (j-1)\cdot h$,
            \If{sum of $\mathcal{W}_{x, y, w, h} \leq \theta$}
            \State $\mathcal{C}[x:x+w, y:y+h] = 0$,
            \EndIf
            \EndFor
        \EndFor
    \end{algorithmic}
\end{algorithm}

\noindent \textit{Line focusing}: To enable  a point-to-point feature matching for laser triangulation,   LLE proceeds to extract a centerline for each laser pattern (generally with a width of greater than 2 pixels) by computing the centroids of the remaining strong laser pattern pixels on a row-by-row basis. By averaging the column indices of the strong curve pixels, the algorithm determines the centroid of the laser line in each row, which are then connected to form a continuous centerline. This focused centerline effectively represents the central path of the laser line in the image, which will be used as the basis for the final continuous curve fitting step discussed next.

\noindent \textit{Curve fitting}: The focused laser line obtained from the last step is not always smooth and continuous. As such, we use a polynomial to fit the rough focused line and thus generate a smooth and continuous curve that accurately represents the laser line in the image. The choice of polynomial order depends on the expected curvature of the laser line on the apple's surface, with higher-order polynomials offering greater flexibility to fit complex shapes. To avoid overfitting problem and through cross-validation,  the $4$th-order polynomial is used in this study to fit curves, which strikes a good tradeoff between accuracy and simplicity based on our preliminary evaluations.

\section{Experiments}
\label{sec:evaluation}

In ALACS, the 3D localization performance is affected by both feature matching accuracy and 3D positions estimation. We thus separately evaluate the performance of LLE and the final apple 3D localization performance. 

\subsection{LLE Evaluation}
\begin{figure*}[!h]
	\centering
	\includegraphics[width=1.00\textwidth]{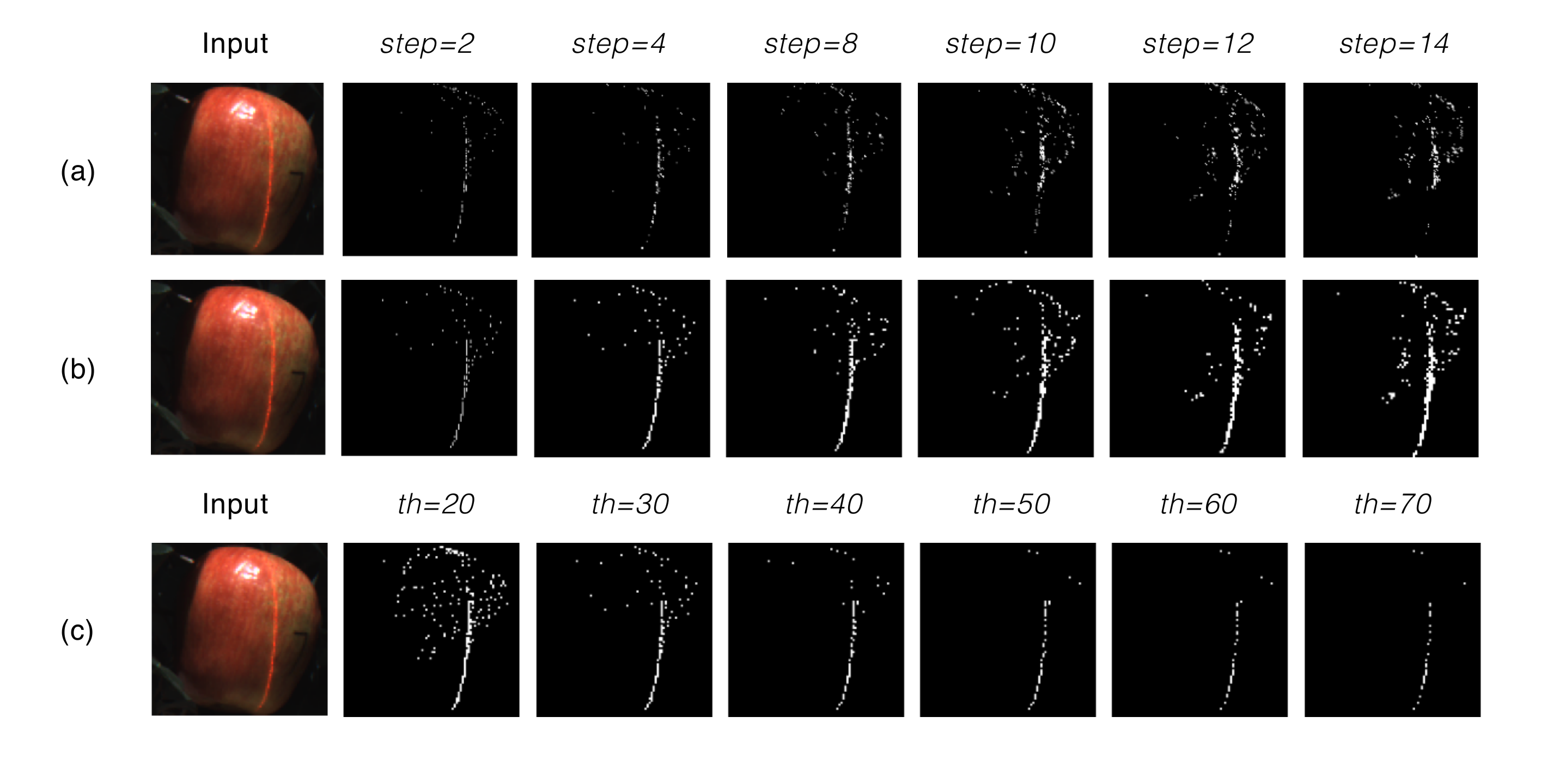}
	\caption{An example of the laser pattern detection performance with different bRCE parameters: (a) with a threshlod value th$=60$ and the step varying from $2$ to $14$; (b) with th$=30$ and the step varying from $2$ to $14$; and (c) with step$=4$ and th varying from $20$ to $70$. The best result is obtained using step$=4$ and th$=40$ for this input.}
	\label{fig:eval_brce}
\end{figure*}

To evaluate the performance of LLE, we conducted a series of experiments under varying lighting conditions (from 1000 to 6500 lux) in the outdoors environment, specifically overcast and  direct lighting scenarios.These tests were intended to assess the robustness and effectiveness of the LLE algorithm in extracting laser lines in challenging environments.
To evaluate the laser line extraction performance using bRCE,  we first varied the bRCE parameters $step$ and $th$ to observe how they affect laser extraction performance (see Figure~\ref{fig:eval_brce}). Based on tests at different distances of $0.8$ m, $1.0$ m, $1.2$ m, $1.4$ m, and $1.6$ m, it was found that the laser lines on the surface of apple are around $4$-pixel wide. When we tuned $step$ from $2$ to $14$ with a fixed $th$, the best results are generated with around step$=4$. With $step$ increases, the shape of laser pattern becomes slacking. A similar approach was used to tune $th$ from $20$ to $70$, and the number of laser pixels tends to decline as  $th$ increases. After calculating the extracted line’s mean and standard deviation from different combination of parameters on $400$ images, the best parameters are chosen to be step$=4$ and th$=40$.

Furthermore, to perform a quantitative evaluation of the LLE algorithm, we calculate the laser line displacements between the LLE predictions and the ground truth, which are obtained through manual labeling. This evaluation provides a quantitative measure of the accuracy and reliability of the LLE algorithm in extracting laser lines under various conditions. 
For this evaluation, a set of images with visible laser lines were manually annotated by trained persons, who carefully trace the laser lines and mark their positions as ground truth. These ground-truth annotations serve as a reference for comparing the performance of the LLE algorithm against an ideal extraction. LLE was then applied to the same set of images, and the resulting laser line predictions were compared with the ground truth annotations. The laser line displacements were calculated as the average pixel-wise Euclidean distances between the points on each row between the LLE-predicted laser lines and the ground-truth laser lines. Smaller displacements indicate a higher degree of agreement between the LLE predictions and the ground truth, reflecting a more accurate and reliable extraction performance. Since we only used central results to localize, we calculated the displacements based on different central segments of each laser line. The results are summarized in Table~\ref{tab:brce}. With calculating displacements in 10\% central segment of the laser line, the LLE generated 1 pixel displacement in average.

\begin{table*}[!ht]
\renewcommand{\arraystretch}{1.2}
\centering
\fontsize{6}{7}
\selectfont
\caption{Performance of the laser line extraction (LLE) algorithm, as measured by average (Avg), minimum (Min) and maximum (Max) displacements (Disp.) in pixels for various  central segment ratios (from $10$\% to $80$\%) on $300$ cases.}
\label{tab:brce}
\resizebox{0.8\textwidth}{!}{
\begin{tabular}{lcccccccc}
\hline
 & 10\% & 20\%   & 30\% & 40\% & 50\%  & 60\% & 70\% & 80\% \\ \hline \hline
 Avg Disp. & 1.0 & 1.2 & 1.3 & 1.4& 1.5& 1.7 & 1.7& 1.9   \\  
 Min Disp. & 0 & 0.2 & 0.3 & 0.4 & 0.4 & 0.5 & 0.5 & 0.6       \\  
 Max Disp. & 2.9 & 3.2 & 3.3 & 3.5 & 3.8 & 3.9 & 4.0 & 4.2     \\ \hline
\end{tabular}}
\end{table*}

\begin{figure*}[!h]
	\centering
	\includegraphics[width=0.95\textwidth]{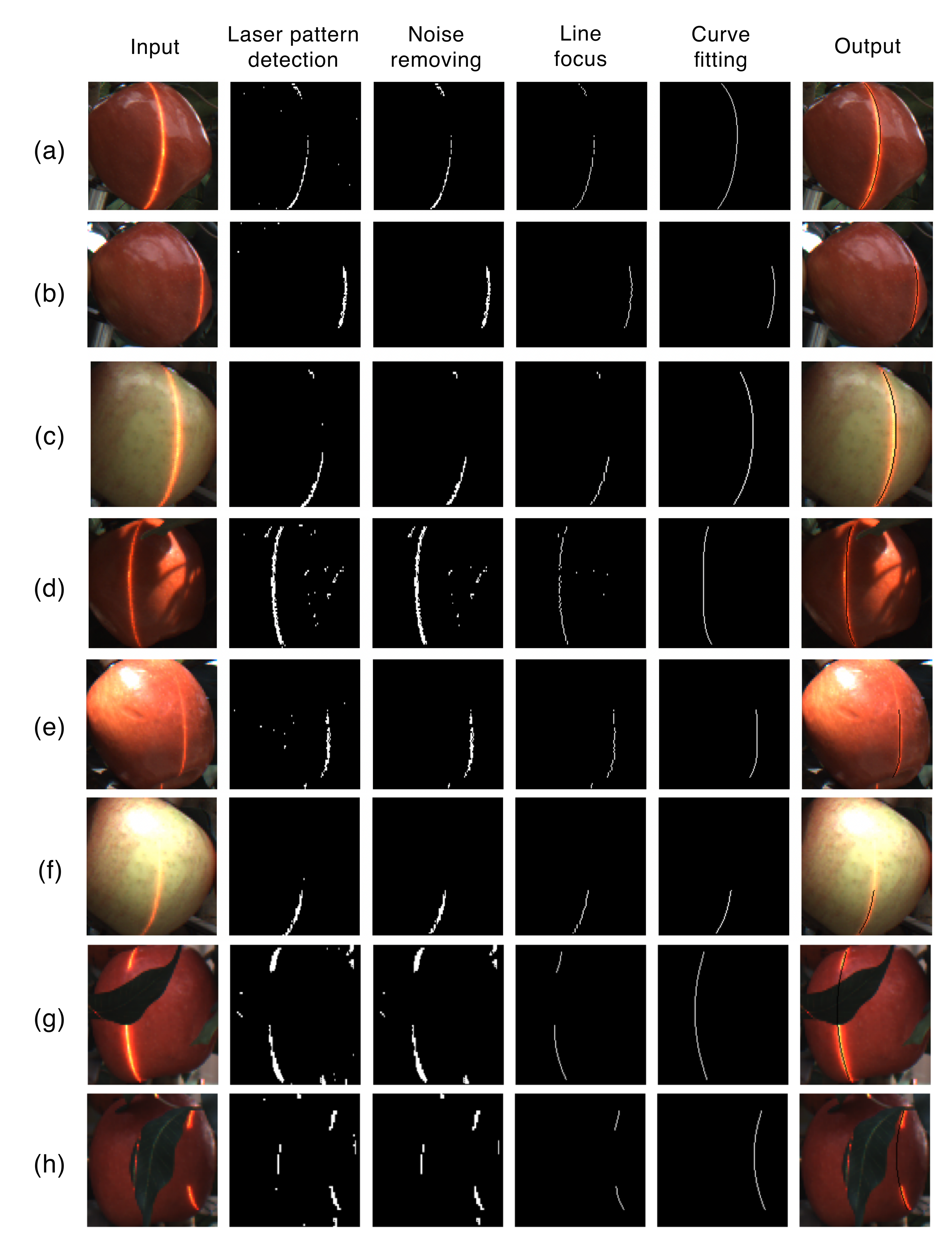}
	\caption{The visualization of the LLE process under different lighting conditions and occlusions (step$=4$, th$=40$): (a)-(c) show the apples with different laser line positions under overcast conditions; (d)-(f) show the apples under direct sunlight; (g) and (h) show the occluded apple cases.}
	\label{fig:brce1}
\end{figure*}

By analyzing the laser line displacements across various images and conditions (see Figure~\ref{fig:brce1}), we can gain insights into the performance and robustness of the LLE algorithm. To avoid image saturation caused by direct lighting, the extracted line by LLE is not always entire and continuous. In the meantime, occlusions also divide laser lines into different parts. These discontinuous challenges are fixed by the polynomial curve fitting (see Figure~\ref{fig:brce1}) and make the final laser line attach to the ground truth. This quantitative evaluation shows us how accurate LLE identify laser patterns even under direct sunlight, ultimately contributing to the overall performance of the ALACS-based apple localization system.

\subsection{ALACS Localization Evaluation}
To assess the performance of the ALACS-based apple localization, we conducted a series of evaluation experiments including occlusion and cluster cases, in both indoor and outdoor environments. In the indoor environment, we compared the localization results obtained by the ALACS method against the ground truth data acquired using a high-precision Qualisys localization system (Qualisys, Sweden) with an accuracy of $0.11$ mm~\citep{carnevale2022virtual}.To facilitate the acquisition of ground truth data, markers were placed on the apples in the orchard, and the Qualisys system was employed to accurately determine their 3D positions. These ground truth positions served as a reference for evaluating the performance of the ALACS-based localization method.

\begin{figure}[!h]
	\centering	\includegraphics[width=0.45\textwidth]{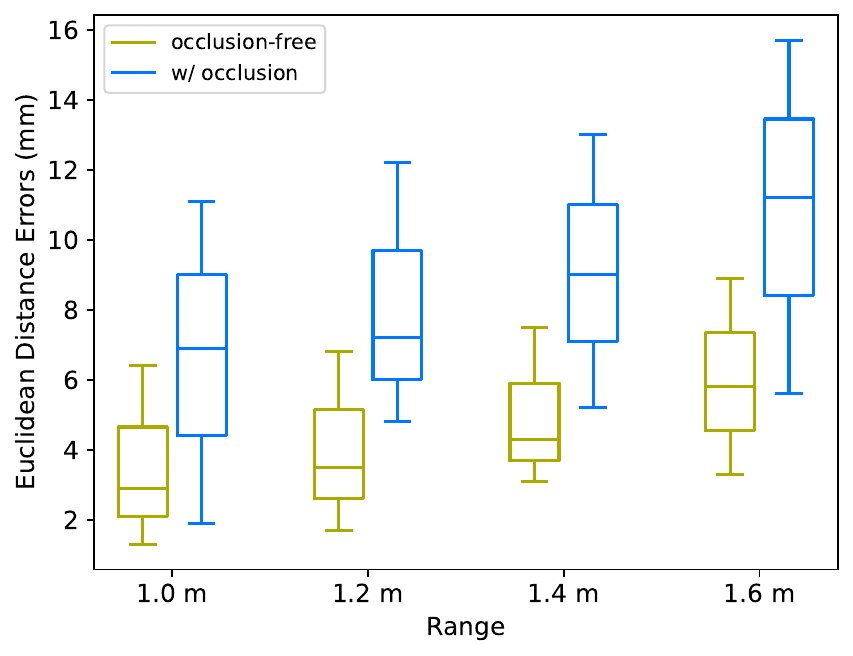}
	\caption{Indoor performance evaluation of the active laser-camera scanning (ALACS) system using single laser projections to the Qualisys marker center from different distances for occlusion-free (120 cases) and occlusion-present (120 cases) situations.}
	\label{fig:single}
\end{figure}

In the first evaluation experiment, we allowed the ALACS system to project a single laser line to the center of the marker placed on the apple, under both occlusion-free and occlusion-present situations. To mimic occlusion situations, we used artificial foliage to partially cover apples. The ALACS system then estimated the apple's position using the extracted laser line and the target position estimation process. The resulting ALACS localization results were compared against the ground truth positions acquired using the Qualisys system. Results in Figure~\ref{fig:single} show that ALACS achieved superior localization performance for occlusion-free situations;  the average distance error ranges from $2.5$ mm to $5.8$ mm at distances from $1.0$ m to $1.6$ m. When the apples were occluded by leaves, the average localization errors were significantly larger than those without occlusions. The average distance errors, under the occlusion situations, are $6.9$ mm, $7.2$ mm, $9.0$ mm and $11.2$ mm respectively at a distance of $1.0$ m, $1.2$ m, $1.4$ m and $1.6$ m. Our harvesting robot uses a vacuum-based end effector to pick fruits, which can tolerate localization errors within $20$ mm~\citep{Zhang2022IROS}. Hence, the ALACS system can still meet the localization accuracy requirements when apples are occluded by leaves.

\begin{figure}[!h]
	\centering
	\includegraphics[width=0.45\textwidth]{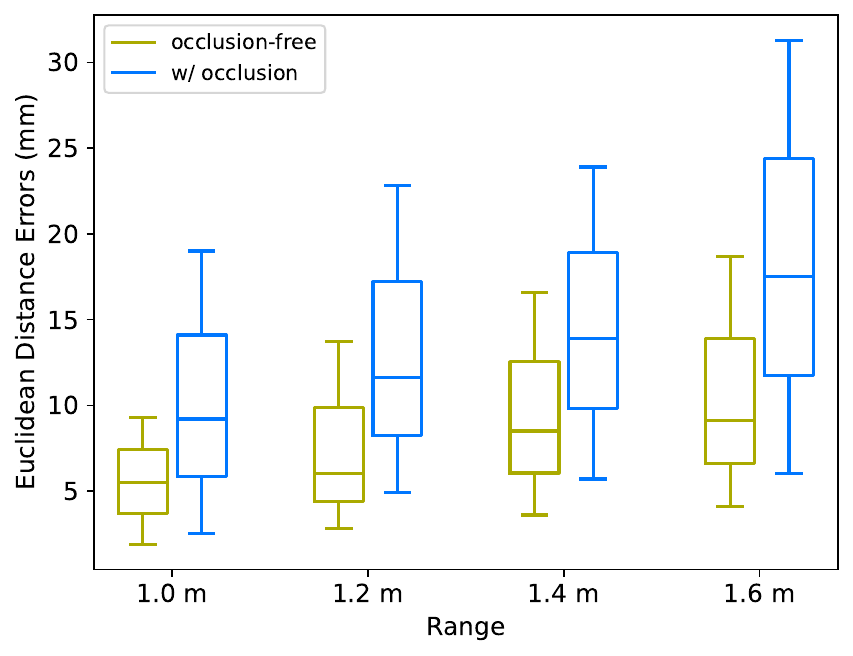}
	\caption{Indoor performance evaluation of the active laser-camera scanning (ALACS) system using multiple laser line projections to the Qualisys marker from different distances under the occlusion-free (120 cases) and occlusion-present (120 cases) situations.}
	\label{fig:multiple}
\end{figure}

In the second indoor evaluation experiment, we tested the ALACS system's ability to estimate the apple's position using multiple laser projections, respectively under occlusion-free and occlusion-present situations. The ALACS system acquired the laser lines at a 2-cm increment for five times over the apple, projecting different laser lines at various positions on the apple's surface. The LLE algorithm was then used to extract the laser lines to determine the apple’s center position based on these five laser projections. As shown in Figure~\ref{fig:multiple}), under the occlusion-free conditions, the average distance error range from $5.5$ mm to $9.1$ mm for distances ranging from $1.0$ m to $1.6$ m, compared with the results obtained in the first experiment for a single laser line for the occlusion free condition. Under the occlusion situation, the average distance errors were $9.2$ mm, $11.6$ mm, $13.9$ mm, and $17.5$ mm at $1.0$ m, $1.2$ m, $1.4$ m, and $1.6$ m, respectively. While these errors are significantly larger compared to those obtained for single laser lines in the first experiment (see Figure~\ref{fig:single}), the ALACS is still expected to  meet our robot localization accuracy requirements of $20$ mm when the distance is less than $1.6$ m, the maximum working distance designed for our harvesting robot~\citep{Zhang2022IROS}.

%By analyzing the evaluation results, it can be found that the shorter distance can provide more accurate positions in the ALACS-based apple localization system.  %This analysis can help identify potential areas for improvement and further optimization of the ALACS system, ultimately contributing to the overall success of the apple localization process in fruit orchards.

Furthermore, we also performed a 3D localization comparison between ALACS and the commercial RGB-D camera RealSense D435i, where the latter is commonly used in the harvesting robots developed by other researchers. In the indoor environment, we still used the Qualysis Motion Tracking System to benchmark the results and test the localization from ALACS with multiple laser projections and RealSense D435i with occlusions at different distances. Table~\ref{tab:compare_indoor} shows the ALACS system significantly outperformed the RealSense benchmark for different distances among $120$ cases.

Since the Qualysis system cannot provide accurate measurements in the outdoor environment due to the varying light condition, we used the positions generated from ALACS and D435i to operate our robotic system~\citep{ZhangJFR2013} to to determine whether fruits would be attached to vacuum-based end-effector. The attachment rate was used as an indirect metric for evaluating the localization results of both ALACS and RealSense D435i. Since our vacuum-based end-effector can tolerate localization errors up to $20$ mm, a measured localization error larger than $20$ mm would be considered a failed detachment. We tested $100$ apples under cloudy and 50\% occlusion rate in a research orchard at Michigan State University’s Horticultural Teaching and Research Center in Holt, MI. Our results showed that ALACS achieved a 95\% detachment rate, whereas the RealSense D435i only had a $71\%$ success rate. 

% \begin{table}[!ht]
% \renewcommand{\arraystretch}{1.2}
% \centering
% \fontsize{6}{7}
% \selectfont
% % Multiple Laser Projections Over the Apple
% \caption{Comparison of sucking rate 
% between ALACS and RealSense D435i over $100$ cases.}
% \label{tab:compare_outdoor}
% \resizebox{0.32\textwidth}{!}{
% \begin{tabular}{lc}
% \hline
%   & Sucking rate   \\ \hline \hline
%  RS D435i & 71\%    \\  
%  \textbf{ALACS} & \textbf{95\%}  \\  
% \hline
% \end{tabular}}
% \end{table}

\begin{table}[!ht]
\renewcommand{\arraystretch}{1.2}
\centering
\fontsize{6}{7}
\selectfont
\caption{Comparison of average localization errors (mm) 
between ALACS and RealSense D435i at different distances.}
\label{tab:compare_indoor}
\resizebox{0.45\textwidth}{!}{
\begin{tabular}{lccccc}
\hline
 Sensor $\setminus$ Range & 1.0m  & 1.2m  & 1.4m & 1.6m   \\ \hline \hline
 RS D435i & 16.0($\pm7.1$) & 17.3($\pm8.0$) & 19.5($\pm9.3$) & 21.5($\pm11.2$) \\  
 \textbf{ALACS} & \textbf{6.9($\pm5.2$)} & \textbf{7.2($\pm5.8$)} & \textbf{9.0}($\pm6.5$) & \textbf{11.2($\pm7.8$)}   \\  
\hline
\end{tabular}}
\end{table}

Results from the three indoor and outdoor experiments have demonstrated that the ALACS system has had significantly enhanced performance for apple localization, compared to RealSense D435i.

\section{Conclusion}
\label{sec:conclu}

In this study, a novel Active Laser-Camera Scanning (ALACS) system was developed for robust apple 3D localization. The proposed LLE method provided precise laser line pattern extractions with an average displacement of $1$ pixel under complex fruit morphology, over-exposure, and occlusion conditions.  For the apple 3D localization, ALACS was able to achieve average errors of $9.2-17.5$ mm at distances ranging from $1.0$m to $1.6$m, which are significantly better, compared to the widely adopted commercial RGB-D camera RealSense D435i. ALACS also demonstrated superior performance for fruit detachment in an apple orchard, when it was tested with our harvesting robot equipped with a vacuum-based end effector. Our future work will include further improvements on the efficiency of ALACS by providing a faster measurement and extending the ALACS coverage width to cover more fruit in our next version.

% \section*{Authorship Contribution}
% \textbf{Pengyu Chu}: Conceptualization, Methodology, Software, Writing – original draft, Writing – review \& editing; \textbf{Zhaojian Li}: Supervision, Resources, Writing - review \& editing; \textbf{Kaixiang Zhang}: Data curation, Writing - review \& editing; \textbf{Kyle Lammers}: Data curation; \textbf{Renfu Lu}: Writing - review \& editing. 

\section*{Acknowledgement}
This research was partially funded by the USDA-ARS inhouse project and National Science Foundation. The authors would also like to acknowledge Michigan State University’s Horticultural Teaching and Research Center in Holt, Michigan and Plant Pathology Research Center in East Lansing, Michigan for their support for collecting the image data in the orchards.

\section*{Disclaimer}

Mention of commercial products is only for providing factual information and does not imply endorsement by USDA over those not mentioned.

\typeout{}
\bibliographystyle{model1-num-names}
\bibliography{manuscript}

\end{document}